%% file: main.tex
\documentclass[sigconf]{acmart}

\setcopyright{none}
\acmConference[DAI '26]{The Eighth International Conference on Distributed
Artificial Intelligence}{November 29--December 2, 2026}{Hong Kong, China}
\acmYear{2026}
\acmBooktitle{The Eighth International Conference on Distributed Artificial
Intelligence (DAI '26), November 29--December 2, 2026, Hong Kong, China}
\copyrightyear{}
\acmDOI{}
\acmISBN{}
\usepackage{booktabs}
\usepackage{amsmath}
\usepackage{xcolor}
\usepackage{graphicx}
\usepackage{float}
\newcommand{\ours}{Operational Validity Contract}
\newcommand{\Rzero}{R_{0}}
\newcommand{\Rone}{R_{1}}

\begin{document}

\title{From Probe Scores to Alarm Policies: Operational Validity of Activation Monitors for Language-Model Agents}

\input{author-information}

\begin{abstract}
Activation probes can predict safety-relevant properties of language models
with high area under the receiver-operating-characteristic curve (AUROC), but
deployed agent monitors make thresholded alarm decisions under tight
false-alarm budgets. These are different estimands. We introduce an
\emph{Operational Validity Contract} that fixes a monitor's target,
observability, identity, timing, intervention unit, comparator, calibration,
and cost. We formalize risk at the semantic request or trajectory level:
when one task contains repeated alarm opportunities, row-level AUROC and false
positive rate do not identify semantic-unit any-alarm risk. A
confidence-certified threshold also requires enough independent negative
units, a tie-safe rule, and transport to deployment.

Across Models Under Pressure, LASR refusal prediction, immutable AgentDojo, and
a prospectively protocol-frozen ST-WebAgentBench replication, joint activation--observable monitors
reach AUROC 0.957 and 0.935 on the first two benchmarks, yet their locked
5\%/10\% detection rates are only .642/.742 and .719/.782, respectively.
On AgentDojo, the secondary mean-activation rollout monitor reaches AUROC 0.922 but detects none
of 38 positive semantic cases at the locked 5\% operating point; thresholds
intended for 10\% false alarms realize 18.7--20.0\% on test. Because the test
misses its prospectively frozen 40-positive support gate, we label it
support-insufficient. On ST-WebAgentBench, activation reaches AUROC .874, but
23 independent calibration negatives cannot identify even a 10\% controller;
the locked policy abstains rather than reporting its mechanical zero FPR as a
success. An exploratory counterexample also lowers full AUROC
while improving realized 10\% utility. The fail-closed compiler caps MUP and
LASR at restricted predictive value and AgentDojo and ST-Web at representation
accessibility; no setting reaches alarm-policy validity. A reproducibility audit
recomputes these claim levels and all four settings' reported score-derived
metrics from frozen predictions and identities.
\end{abstract}


\maketitle

\section{Introduction}

Internal activations are an appealing source of safety signals for
language-model agents. A lightweight probe can reuse a model's forward pass,
operate before a harmful action is completed, and expose features that a small
text classifier may require much more data or compute to recover
\cite{patel2024activation,mckenzie2025highstakes,kirch2025offpolicy}. These
properties have motivated activation monitors for jailbreaks, deception,
refusal, and reward hacking. Their quality is commonly summarized by AUROC,
accuracy, or a gain over a lexical baseline.

An operator, however, does not deploy an AUROC. The operator deploys a score,
an aggregation rule, and a threshold that may stop a trajectory, escalate a
request, or invoke a human. False alarms are paid per semantic task or
trajectory, not necessarily per dataset row. The threshold must be selected
without test labels and must transport across prompts, policies, and tool
environments. A monitor can therefore rank examples well while being unusable
at the false-alarm rate that the alarm policy can afford.

Our AgentDojo experiment makes this gap concrete under a prespecified sampled
retry-set stress audit. We monitor the first policy
request after a Qwen2.5-7B agent consumes an injected tool result and predict
whether each of ten frozen stochastic attempts for the same semantic case
executes the official injection goal. A
frozen mean-activation monitor reaches rollout AUROC 0.9215. At a threshold
locked for a 5\% semantic-case safe-attempt alarm budget, it detects none of
the 38 cases containing an unsafe attempt. At the 10\% budget, the primary
activation and joint monitors realize 18.7\% and 20.0\% safe-attempt alarm
rates. Neither fact is visible from rollout AUROC.
The ten-attempt unit is a stress-audit choice, not an estimate of a
single-execution deployment or a controller that stops at its first alarm.
An independently frozen SuiteCRM replication sharpens the category error:
activation AUROC is .8738, but the prespecified independent-template support
cannot certify a finite threshold, so every 5\%/10\% controller abstains.

We study the resulting question: \emph{when does an activation score constitute
a valid alarm signal for an agent, rather than merely a successful
probe?} Our answer is an \ours{}: Target $\times$ Observability $\times$
Identity $\times$ Timing $\times$ Unit $\times$ Comparator $\times$
Calibration $\times$ Cost. The fields specify the random variable, information
set, generalization boundary, and decision policy to which a result applies.
This is an AgentOps problem: a distributed tool-using system pays intervention
costs over tasks and trajectories, while its monitor is usually validated over
rows. Closing that mismatch is necessary for observable, reproducible, and
governable agent infrastructure.

Our contribution is an activation-monitor-specific integration of established
evaluation and calibration ideas, with three concrete outputs:
\begin{itemize}
    \item a claim-level contract that separates representation accessibility,
    restricted predictive value, and calibrated alarm-policy validity by
    binding the semantic alarm unit, same-horizon comparator, and
    confidence-locked threshold;
    \item a four-setting audit showing that high activation AUROC can coexist
    with poor low-FPR detection, insufficient calibration support,
    calibration-to-test mismatch, and concentrated group burden; and
    \item a fail-closed claim compiler, covered by 20 result-independent
    counterexample tests, and a score-level replay binding identities, timing,
    frozen predictions, thresholds, and immutable test consumption.
\end{itemize}

The calibration inequality, trajectory-level any-alarm risk, and the
distinction between ranking and alarm utility all have prior antecedents. We do
not claim any ingredient separately. Proposition~1 applies the standard
row-to-unit risk relation to this audit rather than proposing a new
concentration inequality. The integration makes explicit which additional
evidence is needed to move from a probe score to an alarm-policy claim.

We do \emph{not} argue that activation monitors universally fail. They improve
low-FPR detection substantially in two settings. The narrower conclusion is
more operationally important: a probe result alone does not identify a valid
alarm policy.

\begin{figure*}[t]
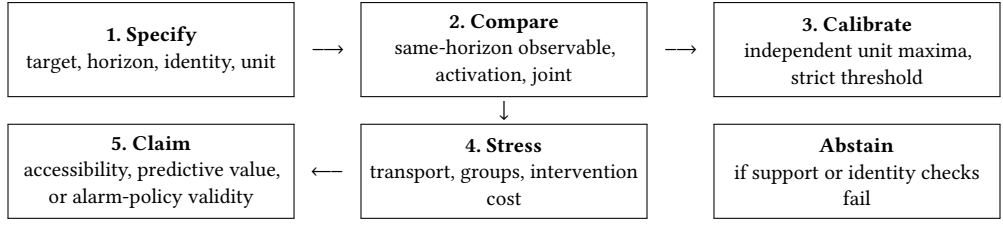

\centering
\setlength{\tabcolsep}{3pt}
\small
\begin{tabular}{ccccc}
\fbox{\parbox[c][1.02cm][c]{0.20\textwidth}{\centering\textbf{1. Specify}\\
target, horizon, identity, unit}}
& $\longrightarrow$ &
\fbox{\parbox[c][1.02cm][c]{0.20\textwidth}{\centering\textbf{2. Compare}\\
same-horizon observable, activation, joint}}
& $\longrightarrow$ &
\fbox{\parbox[c][1.02cm][c]{0.20\textwidth}{\centering\textbf{3. Calibrate}\\
independent unit maxima, strict threshold}}
\\[5pt]
&& $\downarrow$ && \\
\fbox{\parbox[c][1.02cm][c]{0.20\textwidth}{\centering\textbf{5. Claim}\\
accessibility, predictive value, or alarm-policy validity}}
& $\longleftarrow$ &
\fbox{\parbox[c][1.02cm][c]{0.20\textwidth}{\centering\textbf{4. Stress}\\
transport, groups, intervention cost}}
&&
\fbox{\parbox[c][1.02cm][c]{0.20\textwidth}{\centering\textbf{Abstain}\\
if support or identity checks fail}}
\end{tabular}
\caption{The probe-to-policy audit. A high ranking metric is evidence at
Step 2, not permission to skip specification, unit-level calibration, or
transport. Every empirical claim in this paper is labeled by the last step it
passes.}
\Description{Flow diagram from specification through monitor comparison,
calibration, transport stress, and a final claim, with an abstention branch
when identity or support checks fail.}
\label{fig:pipeline}
\end{figure*}

\section{Related Work}

\subsection{Activation monitoring}

Activation monitors have been compared with text classifiers at low FPR,
applied to high-stakes inputs, and tested under off-policy shift
\cite{patel2024activation,mckenzie2025highstakes,kirch2025offpolicy}. Temporal
signals also track refusal and reward-hack risk states
\cite{wilhelm2026reward,collu2026refusal,wilhelm2026riskstates}; monitor
staleness after model updates is a separate transport problem
\cite{duan2026staleness}. These establish accessibility, timing, robustness, or
efficiency, whereas we audit the operational unit and calibrated alarm claim.

AgentLens, TRACES, and AgentForesight move prediction into multi-turn execution
\cite{luo2026agentlens,li2026traces,zhang2026agentforesight}; AcMAS and
PVDetector use activations for multi-agent attacks and policy violations
\cite{xu2026acmas,wang2026pvdetector}. Yet AUC .998 need not identify malicious
content, and a critic with AUROC .94 can reduce success after intervention
\cite{li2026aucnotenough,vasudev2026failureprevention}. General probing controls
raise the same shortcut and comparator concerns
\cite{belinkov2022probing,wang2025falsesense,li2025privileged}. We propose
neither a new detector nor an intervention-effect estimator: our contribution
is the same-horizon observable, semantic alarm unit, threshold-support, and
transport audit between offline prediction and an alarm claim.
Black-to-white performance boosts likewise evaluate internal access relative
to a black-box monitor \cite{parrack2025blackwhite}; a finite-model gain in our
audit remains restricted predictive value, not evidence of privileged or
causal state.

\subsection{Risk-controlled online monitoring}

Conformal risk control calibrates a parameter to bound a monotone expected loss
\cite{angelopoulos2024crc}. Schirmer et al. apply conformal and
high-probability risk control to online LLM verifier signals: one output
sequence is flagged if any step crosses a threshold
\cite{schirmer2026online}. CP-Agent uses held-out trace calibration and a
predeclared finite controller manifest in competitive programming
\cite{wang2026cpagent}. Thus calibrated online monitoring and frozen controller
manifests are not new. Score order statistics and minimum sample-size
requirements for high-probability type-I-error control also predate LLM
monitoring in Neyman--Pearson classification \cite{tong2018nproc}. CORA further
calibrates an execute/abstain boundary for GUI-agent actions
\cite{feng2026cora}. We instead audit whether the score, information set,
nested agent identity, and deployment unit justify an activation-monitor claim.
System-level evaluation already connects errors to Safety Gain, Residual
Hazard, and Availability Cost \cite{guerin2022unifying}; we add a semantic unit,
same-horizon comparison, and finite-sample low-FPR calibration/transport audit,
not a new utility functional.

Low-FPR monitoring is also central to AI control, where a trusted monitor
selects actions or trajectories for audit \cite{greenblatt2024control}. Recent
work models the low-FPR shape of monitor ROC curves
\cite{staley2026logit}. We avoid tail extrapolation and use finite observed
order statistics; unsupported operating points cause abstention.

Three closest contemporaneous systems sharpen our novelty boundary. Monitoring
Monitorability defines intervention, process, and outcome-property archetypes,
a monitorability metric, and a broad agent suite \cite{guan2026monitorability}.
GAVEL composes interpretable activation concepts into configurable real-time
safety rules \cite{rozenfeld2026gavel}. PrefixGuard synthesizes online prefix
monitors across four agent benchmarks and already shows that strong ranking
need not yield actionable first alerts at low false-alarm rates
\cite{huang2026prefixguard}. We do not claim archetype design, operational
activation rules, or the ranking--alarm-utility distinction as new. Our
narrower contribution binds an activation-versus-observable comparison to a
semantic alarm unit, confidence-locked threshold, and frozen transport audit.

General assurance and reporting work provides an additional boundary.
Evaluation Cards composes benchmark, run, and model metadata into an
operational reporting record \cite{ghosh2026evalcards}; structured AI safety
cases bind claims, arguments, and evidence \cite{lee2026safetycases}; and the
Composable Assurance Framework propagates machine-readable safety assertions
through MLOps \cite{zhao2026caf}. The \ours{} is not a general evaluation
schema, safety case, or deployment authorization. It is a narrower statistical
contract for deciding the strongest activation-alarm claim licensed by one
frozen evaluation.

\subsection{Agent safety evaluation}

AgentDojo evaluates prompt injection against stateful tool-using agents using
official utility and security outcomes \cite{debenedetti2024agentdojo}. It is
suitable because the label is executed future behavior rather than a text
proxy. We add stochastic rollouts, a pre-action boundary, semantic-case
grouping, and immutable manifests. SLEIGHT-Bench shows that capable transcript
monitors retain blind spots at 1\% FPR \cite{najt2026sleight}; we do not claim
adaptive robustness. Other benchmarks expose miss/false-alarm trade-offs and
task-disjoint TF--IDF beating LLM judges
\cite{yang2026automonitor,advani2026falsesuccess}. We ask the logically prior
question of whether activations improve a locked alarm policy over such controls.

\begin{table*}[t]
\caption{Scope relative to the closest evaluation and calibration work. A dash
means the item is not the cited work's primary object, not that it is
incompatible with the method.}
\label{tab:closest}
\small
\centering
\resizebox{\textwidth}{!}{%
\begin{tabular}{llllll}
\toprule
Work & Score emphasis & Alarm unit & Same-horizon comparator & Confidence-locked risk & Transport/group audit \\
\midrule
Patel--Wang \cite{patel2024activation} & activation & example & strong text & low-FPR evaluation & adversarial robustness \\
Guerin et al. \cite{guerin2022unifying} & runtime monitor & system outcome & return/hazard model & safety/availability utility & application assumptions \\
Monitoring Monitorability \cite{guan2026monitorability} & CoT/transcript & trajectory/property & action-only & monitorability metric & broad scaling suite \\
GAVEL \cite{rozenfeld2026gavel} & activation rules & rule event & text/guardrail baselines & threshold metrics & domain/rule audit \\
PrefixGuard \cite{huang2026prefixguard} & trace monitor & trajectory first alert & raw text/LLM judges & calibrated FAR points & four-benchmark diagnostics \\
Schirmer et al. \cite{schirmer2026online} & verifier & output sequence & multiple verifiers & sequence any-alarm & exchangeable sequence risk \\
CP-Agent \cite{wang2026cpagent} & trace feedback & program/task & controller variants & held-out risk control & task-specific controller \\
Ours & activation + observable & semantic task/case & declared visible history & unit-max Beta UCB & one-use + suite/support \\
\bottomrule
\end{tabular}
}
\end{table*}

\section{Operational Validity}

\subsection{A monitor claim is a contract}

\paragraph{Definition 1 (contract-bound claim).}
Let a monitor emit score $S$, and let
$\mathcal C=(Y,O,I,H,G,B,P_{\mathrm{cal}},L)$. A validity claim is indexed by
$(S,\mathcal C)$, whose eight fields are:
\begin{enumerate}
    \item \textbf{Target}: an input property, future behavior, completed-output
    property, or causal state;
    \item \textbf{Observability}: the transcript and system information
    available to a non-activation comparator at the same horizon;
    \item \textbf{Identity}: the semantic object that may not cross training,
    calibration, test, or bootstrap clusters;
    \item \textbf{Timing}: when the score is read relative to the action;
    \item \textbf{Unit}: the task, request, rollout, or trajectory on which an
    alarm consumes budget;
    \item \textbf{Comparator}: observable features, model class, data, and
    compute allowed to the baseline;
    \item \textbf{Calibration}: the independent population and rule used to
    lock a threshold; and
    \item \textbf{Cost}: latency, memory, review burden, and intervention loss.
\end{enumerate}
Because $\mathcal C$ selects the random variable, comparator, threshold law,
and loss, the contract fixes an estimand rather than a reporting checklist.

Changing from $\mathcal C$ to $\mathcal C'\neq\mathcal C$ requires an invariance
or transport assumption.

The contract creates a three-level claim ladder. \emph{Representation
accessibility} needs a development-frozen probe and held-out ranking evidence.
\emph{Restricted predictive value} additionally needs a competitive observable
at the same horizon and clustered uncertainty for the incremental gain.
\emph{Alarm-policy validity} additionally needs an operational aggregation rule,
an independently locked threshold, realized risk, sufficient positive support,
transport diagnostics, and declared cost semantics. Evidence cannot be promoted
up this ladder by renaming the target after the experiment. Causal intervention
benefit is audited separately and is not a fourth rung inferred from alarms.

This distinction is invisible to pooled AUROC. With a random positive row
score $S^+$ and negative row score $S^-$,
\begin{equation}
 \mathrm{AUROC}=\Pr(S^+>S^-)+\tfrac{1}{2}\Pr(S^+=S^-).
\end{equation}
It averages all pairwise rankings, has no deployment threshold, weights units
in proportion to their row counts, and assigns no confidence to the extreme
negative tail. It can therefore improve while the alarm policy becomes
worse, as our LASR counterexample later demonstrates.

\subsection{The operational unit}

Let $G$ denote an independent semantic operational unit. It may contain several
rows or stochastic trajectories $i$. Let $Y_{Gi}=1$ denote an unsafe target
event and let larger $S_{Gi}$ be more suspicious. For a prespecified sampled
set of alarm opportunities, define the outcome-conditioned unit maxima
\begin{equation}
 Z_G^- = \max_{i:Y_{Gi}=0} S_{Gi}, \qquad
 Z_G^+ = \max_{i:Y_{Gi}=1} S_{Gi}.
\end{equation}
At threshold $t$, the corresponding operating rates are
\begin{equation}
 \Rzero(t)=\Pr(Z_G^->t), \qquad
 \Rone(t)=\Pr(Z_G^+>t).
\end{equation}
Mixed units contribute to both estimands; repeated rows do not become
independent calibration samples. We call $\Rzero$ the safe-attempt any-alarm
rate and $\Rone$ the unsafe-attempt any-hit rate. They are not the FPR/TPR of a
conventional binary ``any unsafe'' case classifier: a mixed case can contain
both a safe attempt that alarms and an unsafe attempt that is detected.

The meaning of $G$ is part of the operational contract, not a harmless grouping
choice. In AgentDojo our primary $G$ is one semantic security case represented
by a sampled set of ten prespecified attempts. This is a retry-set stress audit
for systems that may retry a task, not an estimate of an ordinary
single-execution deployment or of an online stop controller. We therefore
report fixed-order $k$-attempt and all-single-seed sensitivities separately;
they are different estimands and cannot replace the frozen primary result.
The class-max estimand is evaluated after labeling all sampled attempts; it
does not simulate a sequential controller that halts after its first alarm and
thereby suppresses future retries. Such a stop policy requires a new sequential
contract and prospective evaluation.

\paragraph{Proposition 1 (row risk does not identify unit risk).}
If a safe unit contains $n$ alarm opportunities with identical row FPR $p$,
conditional independence gives
$\Rzero=1-(1-p)^n$. Without a dependence assumption,
$p\leq\Rzero\leq\min(1,np)$. Thus equal row score distributions and equal row
AUROC/FPR can induce different unit risk. At $p=.05$ and $n=10$, unit
FPR is $.401$ under independence. This is an identification error; more
row-level test data cannot repair a misspecified unit.

Schirmer et al. already aggregate any alarm within one output sequence
\cite{schirmer2026online}. Our point concerns the preceding specification:
for an agent, several sequences, seeds, or policy requests can belong to the
same user task on which review burden is paid.

\begin{table}[t]
\caption{A 5\% row false-alarm rate compounds rapidly when a safe operational
unit contains $n$ conditionally independent alarm opportunities.}
\label{tab:compounding}
\small
\centering
\begin{tabular}{lrrrr}
\toprule
$n$ opportunities & 1 & 5 & 10 & 20 \\
\midrule
Unit FPR (\%) & 5.0 & 22.6 & 40.1 & 64.2 \\
\bottomrule
\end{tabular}
\end{table}

\subsection{The alarm-policy objective}

For a false-alarm budget $\alpha$, deployment asks for a feasible threshold
that approximately solves
\begin{equation}
 \max_t\;\Rone(t)\quad\text{subject to}\quad \Rzero(t)\leq\alpha.
\end{equation}
Alternatively, an application can declare benefits and costs and maximize
$b\Rone(t)-c\Rzero(t)-C_{\mathrm{monitor}}(t)$. Both formulations require a
specific threshold and operational unit. Full AUROC integrates over many
thresholds the alarm policy will never use and contains neither intervention
cost nor a risk constraint. We use the constrained form to express a false-alarm
budget; false reviews or blocks motivate this budget, but their
application-specific costs are not measured in these experiments.

A marginal constraint is not a worst-group constraint. If suite $K$ is
observed, a deployment may instead require
$\max_k R_{0,k}(t)\leq\alpha$ or allocate different prespecified budgets by
suite. We report group diagnostics but do not claim such conditional control,
because several AgentDojo suites lack the negative and positive support needed
to calibrate and evaluate it independently.

\subsection{What can an activation add?}

Let $O$ contain all information declared observable at the monitoring horizon,
and let $A=f_\theta(O)$ be a frozen deterministic activation.

\paragraph{Proposition 2 (information ceiling).}
$I(Y;A\mid O)=0$, and under an unrestricted decision class the Bayes risk using
$O$ is no worse than that using $A$ alone. Conditioning on $O$ fixes $A$, and
any rule $h(A)$ can be written $h(f_\theta(O))$.

This does not make finite activation probes useless. Held-out probe performance
can establish representation accessibility; an incremental held-out gain over
the declared observable comparator can establish restricted predictive value.
Sample or compute efficiency additionally requires a learning curve or
resource-matched measurement. None implies privileged information or hidden
intent. If $O$
omits information used to compute $A$, the claim must state that omission
explicitly.

\subsection{Certified thresholds, support, and ties}

Conditional on the frozen score function, suppose $m$ calibration samples of
$Z_G^-$ are independent draws from a common negative-unit distribution,
also shared by an independent deployment unit. Sort them descending,
and $T_r$ is the $r$th largest. We alarm iff $Z_G^->T_r$. For any score
distribution, including discrete or
quantized scores, the new-unit strict tail mass
$q(T_r)=\Pr(Z_{\mathrm{new}}^->T_r\mid T_r)$ is stochastically upper-bounded by
$\mathrm{Beta}(r,m-r+1)$; under continuity the bound is exact. Therefore
\begin{equation}
 \Pr_{\mathrm{cal}}\!\left[q(T_r)\leq
 B^{-1}_{r,m-r+1}(1-\delta)\right]\geq 1-\delta.
\end{equation}
We lock the largest rank $r$ whose upper quantile is at most the false-alarm
budget $\alpha$. If no rank qualifies, the alarm policy abstains. At rank one, a
finite threshold requires
\begin{equation}
 m \geq \frac{\log\delta}{\log(1-\alpha)}.
\end{equation}
For $\alpha=\delta=.05$, this is 59 independent negative units merely to
identify a finite threshold; it says nothing about TPR. Selected-value ties
make the bound conservative rather than invalid under strict exceedance. The
positive-test minima used below are frozen design-sufficiency gates, not
consequences of this negative-tail inequality. The
artifact records selected-value multiplicity, calibration strict exceedances,
and test equalities; all reported test operating points have zero equality.
The pre-test AgentDojo threshold JSON and earlier replay recorded the legacy
non-strict comparison. The submission audit changes only that comparator to
strict exceedance: numeric thresholds, ranks, and every test decision remain
unchanged. We expose this post-hoc semantics correction rather than treating
sample uniqueness as proof of population continuity.

This analysis separates three bottlenecks. \emph{Tail identifiability} asks
whether calibration support can certify a threshold. \emph{Tail separability}
asks whether any valid threshold has useful $\Rone$. \emph{Transport} asks
whether the calibration and deployment unit distributions match. More
calibration data can improve the first, but not guarantee the other two.

The distinction also clarifies what a confidence statement means. The
order-statistic bound controls the marginal probability that a new
independent negative unit from that same distribution strictly exceeds the
selected score. Mere exchangeability without independence is insufficient
for this high-probability population-risk bound. It is not a
simultaneous guarantee over suites,
families, policies, or an adaptive sequence of redeployments. Those require
additional multiplicity control or a new calibration contract.

\section{Evaluation Protocol}

\subsection{Four targets}

The four settings are case studies of a common audit procedure, not a
controlled cross-benchmark comparison or a sample from a single deployment
population. Each asks which claim its own target, unit, comparator, and
calibration evidence support. We report within-setting comparisons without
pooling effect sizes or estimating how often activation monitors succeed or
fail in deployment.

Table~\ref{tab:claims} records the claim before any operating-point comparison.
MUP contains independent high-stakes inputs and tests an observable input
property \cite{mckenzie2025highstakes}. LASR predicts whether Llama-3.2-3B will
refuse before any assistant token, using recovered original requests as units
\cite{kirch2025offpolicy}. These analyses use frozen splits but are
retrospective because earlier benchmark results had been inspected.

The released LASR row split is not a pure semantic OOD split: 207 of 969 test
request identities occur in the official training set, including 52 that occur
in our frozen calibration subset. We therefore label official-all results as a
published-split description, not independent transport evidence, and report
the protocol-declared semantic-unseen subset (762 requests) as the identity-
disjoint sensitivity. The subset is defined only by request identity, without
using labels or monitor scores.

AgentDojo is the prospective future-behavior test. We freeze AgentDojo v1,
Qwen2.5-7B-Instruct, temperature .7, top-$p$ .9, three then ten seeds for model
fitting, and disjoint user-task splits. The monitoring boundary is the first
policy request after an injected tool result. The label is whether the
remaining trajectory executes the official injection task. A global constant
marker replaces an early identity-bearing marker before any monitor feature is
collected. One immutable 1,190-identity test manifest, containing ten
prespecified stochastic attempts per semantic case, is consumed once; failed
trajectories are retained without regeneration. This is not the same estimand
as observing one trajectory per case.

ST-WebAgentBench is a separately protocol-frozen prospective replication on SuiteCRM
\cite{levy2026stweb}. Its unit is an intent template with safe and unsafe
trajectory siblings; the monitor reads Qwen2.5-7B layer 14 before each browser
action. Of 192 immutable test identities, 173 valid rollouts yield 2,237
pre-action rows. Twelve engineering failures are retained without replacement.
The exhaustive ledger assigns all result-bearing input, future-response,
tool-agent, and browser-agent branches to this table; it is not a random task
sample, and rejected, failed, and negative branches remain in the supplement.

\begin{table}[t]
\caption{Frozen claim cards. ``Pre'' means before the predicted behavior.}
\label{tab:claims}
\small
\begin{tabular}{llll}
\toprule
Setting & Target & Timing & Unit \\
\midrule
MUP & high-stakes input & input & input \\
LASR & future refusal & pre-response & semantic request \\
AgentDojo & injection execution & pre-action & security case \\
ST-Web & task failure & pre-action & intent template \\
\bottomrule
\end{tabular}
\end{table}

For AgentDojo, the eight fields instantiate as: executed injection goal
($Y$); exact visible history through the injected tool result ($O$); user task
and frozen model/policy identity ($I$); first subsequent policy request ($H$);
the sampled ten-attempt security case ($G$); TF--IDF, tool/suite metadata, and
three Qwen-14B forecasts ($B$); disjoint negative-case maxima with the frozen
strict order statistic ($P_{\mathrm{cal}}$); and safe-attempt alarm burden
versus unsafe-attempt hits ($L$). Downstream intervention cost is not measured,
so this audits risk proxies rather than net deployment utility.

\subsection{Monitor ladder}

Each setting compares observable, activation, and joint monitors. MUP uses a
frozen lexical observable and a validation-selected probe. LASR's observable
combines lexical features with three Qwen-14B semantic forecasts
available from the same prompt. Its activation is the corrected layer-12
cumulative mean. AgentDojo's observable combines canonical visible-history
character/word TF--IDF, suite/tool metadata, and three frozen Qwen-14B
future-outcome log-odds prompts. The primary activation is Qwen2.5-7B
zero-indexed decoder layer 14 at the final token of the exact policy request;
full-request mean is a prespecified secondary analysis. Late fusion uses
out-of-fold base predictions with grouped folds. No AgentDojo layer, pooling,
template, feature, solver, or threshold is chosen from calibration or test
performance.
ST-Web reuses the frozen layer/pooling and strong-observer construction, with
intent-template-grouped fitting and no outcome-dependent revision.

The declared observer is intentionally capable where public artifacts permit:
does activation add predictive utility beyond what the same visible history
can support? The settings nevertheless use heterogeneous observer classes
(lexical for MUP; Qwen-14B-assisted for LASR and AgentDojo), so we make
within-setting increments, not cross-setting comparator rankings. Compute is
reported separately rather than assumed equal.

\subsection{Calibration and integrity}

Model fitting and threshold calibration use disjoint semantic units. We lock
95\%-confidence marginal thresholds for $\alpha\in\{.05,.10\}$ from negative
unit maxima. Results report realized FPR/TPR, paired 5,000-sample semantic-unit
bootstrap intervals, and suite/family diagnostics. Each budget binds its own
minimum TPR; all four contracts froze .80 before operating-point analyses. It
is not a universal optimum or a hypothesis-test threshold; the
scientific conclusions use the reported rates, intervals, support, and
transport audits rather than this binary rule.

Coverage additionally requires independent calibration and deployment units
from the same negative-unit distribution. Because the official LASR test contains the disclosed training
identity overlap, we attach no transport certificate to that all-test panel;
the semantic-unseen panel is the corresponding identity-disjoint check. MUP,
AgentDojo, and ST-Web use their declared independent test units.

For a mixed case, safe and unsafe stochastic outcomes enter $Z_G^-$ and
$Z_G^+$ respectively: the same user task can create both a safe-attempt alarm
and an unsafe-attempt detection opportunity. Paired differences reuse each resampled
semantic case across monitors; percentile intervals are computed from 5,000
case-cluster bootstrap replicates. The calibration UCB and the realized test
FPR answer different questions, so neither is substituted for the other. We
treat realized FPR as a transport diagnostic, not proof of exchangeability; a
point estimate below $\alpha$ cannot itself certify deployment shift.
MUP intervals resample medical specialties and LASR intervals resample
semantic requests; AgentDojo intervals resample semantic security cases, and
ST-Web intervals resample intent templates.
Across monitor, budget, and sensitivity panels we report pointwise intervals
without familywise multiplicity control. We do not select a significant cell
or reinterpret these intervals as simultaneous guarantees.

\begin{figure}[t]
\centering
\fbox{\begin{minipage}{0.93\columnwidth}
\small
\textbf{Frozen alarm-policy audit procedure}
\begin{enumerate}
\setlength{\itemsep}{1pt}
\setlength{\parskip}{0pt}
\item Declare target, visible history, semantic identity, horizon, unit, score
families, and promotion rule.
\item Split by semantic identity; fit Observable and Activation only on model
fit, and fit Joint from grouped out-of-fold base predictions.
\item Freeze model and feature hashes. On calibration only, form negative unit
maxima and select the largest Beta-certified rank.
\item Apply strict exceedance, audit support, selected-value multiplicity, and
test equalities; abstain only when no rank has the required upper bound.
\item Hash the thresholds, consume test once, retain failures, and report unit
FPR/TPR, paired monitor differences, and group support.
\end{enumerate}
\end{minipage}}
\caption{No layer, pooling rule, template, model, or threshold is changed after
its corresponding freeze point.}
\Description{Five-step alarm-policy audit: declare the estimand, fit grouped
monitor models, freeze hashes, calibrate unit maxima with tie checks, then
consume test once and report operational and group metrics.}
\label{fig:algorithm}
\end{figure}

The claim compiler applies the claim ladder in Section~3 to frozen evidence;
it never fits or searches. Appendix~F specifies its decision rule. Twenty
result-independent counterexamples cover malformed identities, budgets, ranks,
ties, support, mechanical zeros, and promotion. A separate replay reconstructs
all four settings' score metrics from allowlisted frozen archives; upstream
audits separately verify prompt/token and activation alignment, hashes, and
immutable test consumption.

\begin{table}[t]
\caption{Artifact checks that prevent a ranking result from silently changing
its estimand.}
\label{tab:audit}
\small
\begin{tabular}{p{0.19\columnwidth}p{0.72\columnwidth}}
\toprule
Boundary & Immutable check \\
\midrule
Identity & semantic hashes; grouped folds; overlap/non-overlap audit \\
Timing & exact visible prefix; tool-result boundary reconstruction \\
Features & layer, pooling, templates, payload and model hashes \\
Calibration & negative-unit count, selected rank, Beta UCB, strict rule/ties \\
Test & one-use manifest, failed-rollout retention, no refitting \\
\bottomrule
\end{tabular}
\end{table}

\subsection{Resource accounting}

Predictive dominance and deployment preference are different comparisons. In
a separate hook microbenchmark on the AgentDojo Qwen-7B stack, capturing the
frozen layer added no resolvable batch-one latency and 0.067\% at batch eight;
the captured tensor occupied 4--41 MiB across tested sequence lengths, against
an approximately 26 ms batch-one forward pass. These measurements cover
activation extraction, not the downstream probe or the Qwen-14B observer. We
therefore report the latter as a strong information-set comparator, not as a
compute-matched replacement. A cheap activation can remain useful in a cascade
even when it has no independent predictive increment. Human-review and action-
blocking costs are unmeasured, so we compare risk proxies and do not claim net
deployment utility.

\section{Results}

We organize evidence by the three gates induced by the contract rather than by
benchmark: whether a low-FPR threshold is identifiable, whether the certified
tail is separable, and whether calibration transports. Retry and suite analyses
then test whether the declared unit hides policy or group heterogeneity.

\subsection{Calibration support and threshold identifiability}

The first three settings contain enough independent negative calibration units to
select finite thresholds at both budgets. Table~\ref{tab:calibration} reports
the frozen rank and its 95\% upper confidence bound (UCB). This rules out the
simplest explanation---that poor utility is only an undefined percentile from
too few negative rows. It does not establish positive-class separation or
transport. AgentDojo illustrates the distinction: its 10\% activation threshold
has an 8.92\% calibration UCB but realizes 18.67\% test FPR. Under the
continuous independent-and-identically-distributed null, integrating over the frozen rank-5 threshold
gives $K\sim\mathrm{BetaBinomial}(75,5,96)$ and
$\Pr(K\geq14)=.001745$; strict exceedance is conservative when scores tie.
This predictive check includes calibration-threshold uncertainty and diagnoses
a calibration-to-test distribution mismatch, not an intervention effect.
ST-Web exposes the logically prior failure. Its 23 calibration-negative intent
templates are below the rank-one requirements of 59 at 5\% and 29 at 10\%.
All frozen thresholds are therefore $+\infty$. The resulting zero alarms are
mechanical abstention, not successful transport.

\begin{table}[t]
\caption{Independent negative-unit support and realized primary-activation
FPR. Selected ranks depend on $m$, $\alpha$, and $\delta=.05$, not test
performance.}
\label{tab:calibration}
\scriptsize
\centering
\begin{tabular}{lrrrrr}
\toprule
 & $m$ & \multicolumn{2}{c}{5\% budget} & \multicolumn{2}{c}{10\% budget} \\
Setting &  & $r$/UCB & Test FPR & $r$/UCB & Test FPR \\
\midrule
MUP & 69 & 1/.0425 & .0099 & 3/.0884 & .0199 \\
LASR & 372 & 12/.0485 & .0183 & 28/.0987 & .0872 \\
AgentDojo & 100 & 2/.0466 & .0000 & 5/.0892 & .1867 \\
ST-Web & 23 & \multicolumn{2}{c}{abstain} & \multicolumn{2}{c}{abstain} \\
\bottomrule
\end{tabular}
\end{table}

\subsection{High AUROC does not ensure low-FPR detection}

Table~\ref{tab:main} reports the joint monitor, which is the most favorable
summary for an activation increment on MUP and LASR. MUP reaches AUROC .9571,
yet its realized 5\% and 10\% TPRs at FPR-certified thresholds are .6424 and
.7417. LASR reaches request-event AUROC .9348; the corresponding realized TPRs
are .7189 and .7821. Both improve over
observable baselines, so the result is not ``activations fail.'' Neither meets
the frozen requirement of at least .80 TPR at both budgets.

The conclusion persists on 762 identity-disjoint, semantic-unseen LASR
requests. AUROCs are .9066 (Observable), .9424 (Activation), and .9427 (Joint).
Their 5\% FPR/TPR pairs are .0334\slash.6234, .0180\slash.7065, and
.0206\slash.7065; at 10\% they are .1054\slash.7403, .0643\slash.7662, and
.0720\slash.7740.
Joint-minus-Observable TPR is +.0831,
request-bootstrap 95\% CI $[.0494,.1169]$, at 5\%, and +.0338, CI
$[.0052,.0623]$, at 10\%. Thus overlap does not create the observed increment,
but the unseen alarm policy still misses the frozen .80 utility gate.

Paired cluster uncertainty also distinguishes the two budgets. MUP
Joint-minus-Observable TPR is +.2649, specialty-bootstrap 95\% CI
$[.1529,.4268]$, at 5\%, but +.0762, CI $[-.0192,.2209]$, at 10\%. LASR's
request-bootstrap increments are +.1059, CI $[.0754,.1385]$, and +.0407, CI
$[.0122,.0692]$, respectively. Thus activation can materially improve a
restricted comparator while the locked alarm policy still misses its utility
requirement.

\begin{table*}[t]
\caption{Locked operational-unit results. AUROC is test unit-level AUROC except
AgentDojo, where both primary case-max and the highest prespecified rollout
AUROC are shown. FPR/TPR are realized at independently calibrated thresholds.
LASR rows show the official all-test split; the identity-disjoint sensitivity
is reported in text and Appendix~B, Table~\ref{tab:lasr-unseen}. ST-Web dashes denote
prespecified abstention, not observed zeros. ADJ = AgentDojo.}
\label{tab:main}
\small
\begin{tabular}{llrrrrr}
\toprule
Setting & Monitor & AUROC & 5\% FPR & 5\% TPR & 10\% FPR & 10\% TPR \\
\midrule
MUP & Observable & .9396 & .0033 & .3775 & .0497 & .6656 \\
       & Activation & .9446 & .0099 & .5000 & .0199 & .6457 \\
       & Joint & \textbf{.9571} & .0066 & \textbf{.6424} & .0232 & \textbf{.7417} \\
\midrule
LASR & Observable & .8987 & .0467 & .6130 & .1176 & .7413 \\
       & Activation & .9344 & .0183 & \textbf{.7189} & .0872 & .7780 \\
       & Joint & \textbf{.9348} & .0243 & \textbf{.7189} & .0913 & \textbf{.7821} \\
\midrule
ADJ & Observable & .7860 & .0267 & .0526 & .2000 & \textbf{.5789} \\
       & Activation last & \textbf{.8028} & .0000 & \textbf{.0526} & .1867 & .5263 \\
       & Joint last & .7986 & .0000 & .0263 & .2000 & \textbf{.5789} \\
ADJ rollout & Activation mean & \textbf{.9215} & .0000 & .0000 & .0000 & .1053 \\
\midrule
ST-Web & Observable & .8497 & -- & -- & -- & -- \\
       & Activation last & \textbf{.8738} & -- & -- & -- & -- \\
       & Joint last & .8712 & -- & -- & -- & -- \\
\bottomrule
\end{tabular}
\end{table*}

Appendix Figure~\ref{fig:operating-points} visualizes the AUROC--operating-point
mismatch without pairing row-level discrimination with unit-level TPR.

The LASR nominal, non-certified 10\% activation threshold reaches .8126 TPR,
but realizes .1136 FPR. At 5\%, even its nominal TPR is .7475. Thus LASR's 10\%
near miss is partly calibration-resolution limited, whereas its 5\% failure is
primarily tail separation. A fixed-data allocation sweep later increased
calibration support without rescuing the operating point, illustrating that
support and separation cannot be substituted mechanically.

\subsection{AgentDojo calibration-to-test mismatch}

\paragraph{No model selection on test.}
Before calibration or test behavior was generated, grouped model-fit
out-of-fold AUROC was .9020 for Observable, .8918 for Activation last, and .8879
for Joint last. The Joint-minus-Observable difference was $-.0141$ with paired-case
95\% CI $[-.0338,.0041]$. On mixed cases it was $-.0469$ with CI
$[-.0990,-.0039]$. These diagnostics argued \emph{against} an activation
increment. We nevertheless locked the declared pipelines and consumed the
test manifest once. This negative pre-test evidence is important: the reported
test is a falsification attempt, not a selected success from layers or pooling
rules.

The AgentDojo test contains 1,190 frozen rollout identities, of which 1,165
complete. There are 602 eligible monitoring-boundary rows across 82 semantic
cases, with 75 negative cases, 38 positive cases, and 31 mixed cases. Trace
alignment and exact prompt reconstruction are 100\%. The prospectively frozen minimum
was 40 positive cases, so the formal status is \emph{support-insufficient}; this
was a frozen design-sufficiency rule distinct from the analytic negative-tail
certificate, and we do not add or rerun cases. The primary increment and 10\%
transport conclusions below fail independently of this two-case shortfall.

The primary case maximum corresponds to the sampled ten-attempt retry-set
audit, not an intrinsic property of ``AgentDojo.'' Prespecified fixed-order
$k=1,3,5,10$ and single-seed sensitivities (Appendix~D) vary sharply; only four single-seed policies
identify a finite 5\% threshold. They do not replace the
primary estimand, but show why retry semantics must be declared in advance.

At 5\%, Observable detects two positive cases with two false alarms. Activation
last detects the same number with no false alarms, while Joint last detects one
with none. The primary Joint-minus-Observable TPR difference is $-.0263$, paired
semantic-case bootstrap 95\% CI $[-.0789,0]$. It fails the promotion rule before
considering the support shortfall. At 10\%, Observable and Joint make exactly the
same positive-case decisions, but both realize .20 FPR. The activation-only FPR
is .1867 and TPR .5263.

Pooling reinforces the category error. Mean activation has the best rollout
AUROC in Table~\ref{tab:main}, yet zero 5\% TPR and only .1053 TPR at 10\%.
Because mean pooling was frozen as secondary, this is a sensitivity rather than
a post-hoc model switch.

\subsection{The declared unit exposes suite concentration}

AgentDojo alarms concentrate by suite: at 5\%, all true positives and both
Observable false positives occur in Slack; its 14.29\% Slack FPR contrasts
with 2.67\% pooled FPR. Sparse suite support precludes certificates but rejects
interpreting a marginal rate as uniform protection.

The failure modes are therefore not interchangeable. MUP has modest positive
tail overlap despite conservative realized FPR. LASR has adequate support and
passes the 10\% target only under the exploratory tail-aligned objective, but
its 5\% tail remains poorly separated. AgentDojo has formal negative-tail
support but exhibits a calibration-to-test mismatch at 10\%, has negligible
5\% positive recall, and
exhibits severe suite concentration. ST-Web is earlier still: its activation
AUROC is .8738 and Joint-minus-Observable AUROC is +.0215, template-bootstrap
95\% CI $[-.0097,.0568]$, but calibration support makes both controllers
unidentifiable; its 28 negative test templates also miss the prospectively frozen 29.
Reporting only ``did not reach 80\% TPR''
would erase the scientific distinction among identifiability, separation, and
transport.

The compiler therefore caps MUP and LASR at \emph{restricted predictive value}:
their comparator increments are supported, but their retrospective contracts
did not prespecify test-support gates. AgentDojo and ST-Web stop at
\emph{representation accessibility} because their frozen comparator increments
fail; support and transport failures independently preclude alarm-policy
validity. No current setting licenses an intervention-effect claim.

\subsection{Objective mismatch: AUROC and utility can move oppositely}

An exploratory LASR unit-max partial-AUC objective over FPR $[0,.10]$ lowers
AUROC .9344$\rightarrow$.9184 but raises certified 10\% TPR
.7780$\rightarrow$.8106 while FPR falls .0872$\rightarrow$.0669. This
constructively reverses full-AUROC selection. Partial-AUC is prior art
\cite{narasimhan2013pauc}, test results were already inspected, and a separately
frozen 5\% variant fails (.7189$\rightarrow$.5743 TPR); this is a diagnostic,
not a promoted optimizer.

\section{Discussion}

\subsection{Implications for AgentOps}

Monitor registries should bind every score to the eight contract fields and
reject identity leakage, row/unit mismatch, unsupported ranks, ties, or drift.
Dashboards should report unit-level rates and abstain on unidentified policies.

Activation access remains useful under this standard. On MUP and LASR it
materially improves low-FPR recall over the declared restricted observable
models. Cheap probes may form a cascade, but without a learning curve or
resource match, predictive increments do not establish privileged information,
efficiency, or deployment validity.

AgentDojo also shows that suite or policy shift can violate exchangeability
even with immutable code and thresholds.

\subsection{What the evidence does and does not establish}

The cross-setting pattern supports accessibility and, retrospectively on
MUP/LASR, improved low-FPR recall over restricted observables. LASR additionally shows, only
exploratorily, that an operational objective can beat full-AUROC selection.
Neither result implies privileged information, causality, or transport.

Alarm-policy evaluation therefore requires a declared unit and horizon, a
capable same-horizon comparator, independent-unit calibration, and distinct
support, separation, and transport tests. Each can change the conclusion.

\subsection{Limitations}

AgentDojo and ST-Web are independently frozen prospective evaluations; both
retain failures and miss a prespecified support gate. MUP and LASR are
retrospective. Both prospective tests study one Qwen-7B layer and frozen
pooling; AgentDojo covers one attack and four sparse suites, while ST-Web is
SuiteCRM-only. Certificates are marginal, not adversarial or familywise.
The Qwen-14B comparators are not compute matched, so their contrasts establish
only predictive differences under the declared information sets, not relative
cost-effectiveness. We do not measure human-review burden, blocking cost,
intervention benefit, or net deployment utility. The contract records these
missing quantities explicitly; an audit verdict is not a complete deployment
assessment.

AgentDojo's sampled ten-attempt unit is policy-specific; post-hoc sensitivities
cannot choose deployment semantics. A single-execution or stop-on-alarm policy
would require its own calibration and prospective evaluation. The exhaustive screening ledger is not a
random task sample, so the audit does not estimate failure prevalence.

The four heterogeneous targets form a validity audit, not a pooled effect.
In particular, the prospective studies demonstrate rejection of unsupported
alarm claims rather than validate a successful alarm policy; their limited
support cannot establish a general failure of activation monitoring.
Qwen-14B forecasts are automated comparators, not human judgments. Adaptive
attacks and calibration drift remain open.

The artifact verifies frozen scores, identities, contracts, and paper numbers;
it does not regenerate behavior/features or refit, so reproducibility is
score-level rather than end-to-end.

\subsection{Ethical and societal considerations}

The public experiments use model-generated trajectories. Invalid thresholds
can concentrate burden; unsupported policies should abstain and use group
reporting, least privilege, and escalation. Logs require access control and
minimal retention; we release no new exploit payloads.

\section{Conclusion}

A probe asks what is accessible; an alarm policy asks whether a locked decision
meets operational risk and utility requirements. Across four targets, semantic
units, same-horizon observables, calibration, support, and transport change the
answer suggested by AUROC. The contribution is their fail-closed,
activation-alarm-specific binding---not a new detector or calibration theorem.
Deployment validity is a separate claim.

\section*{Generative AI Disclosure}

Generative-AI tools assisted with language editing and draft code generation.
The author independently checked the research design, source literature,
mathematical claims, code, artifact hashes, and every reported numerical result.

\label{body-end}
\bibliographystyle{ACM-Reference-Format}
\bibliography{references}

\clearpage
\appendix
\input{appendix}

\end{document}

%% file: author-information.tex
\author{Xueping Gao}
\affiliation{%
  \institution{Alibaba Cloud Computing}
  \city{Hangzhou}
  \country{China}
}
\email{xueping.gxp@alibaba-inc.com}

%% file: appendix.tex
\begin{figure*}[!t]
\centering
\includegraphics[width=0.91\textwidth]{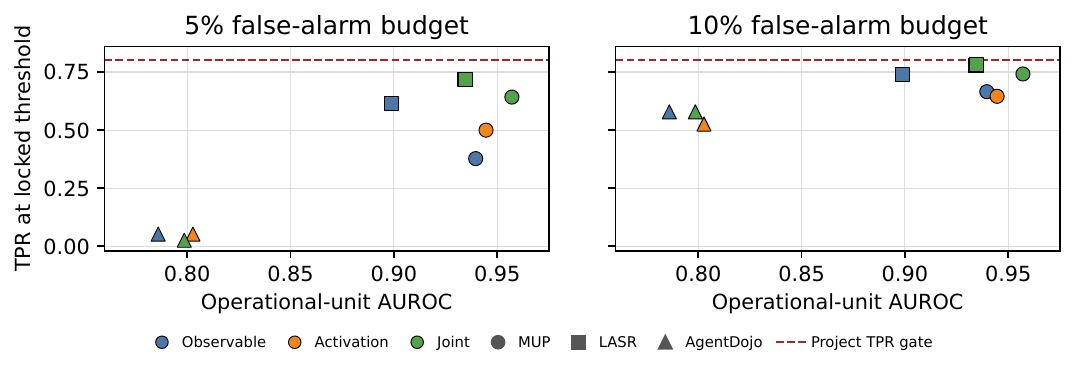}
\caption{Operational-unit AUROC versus TPR at each independently locked
threshold. Marker shape identifies the setting and color the monitor. The
dashed .80 line is a frozen project promotion rule, not a universal utility
threshold; all realized rates are reported. The secondary AgentDojo rollout
AUROC is excluded because pairing rollout AUROC with case-level TPR would mix
estimands; ST-Web is excluded because no finite controller was identifiable.}
\Description{Scatter plot comparing operational-unit AUROC and realized TPR
at 5 and 10 percent false-alarm budgets for observable, activation, and joint
monitors on MUP, LASR, and AgentDojo. ST-Web is absent because its frozen
policy abstains.}
\label{fig:operating-points}
\end{figure*}

\section{Proof Sketches and Calibration Details}

\subsection{Operational-unit compounding}

Let $E_i$ be the event that the $i$th opportunity in a safe unit alarms, with
$\Pr(E_i)=p$. If the events are conditionally independent,
\begin{equation}
 \Pr\left(\bigcup_{i=1}^{n}E_i\right)
 =1-\Pr\left(\bigcap_{i=1}^{n}E_i^c\right)=1-(1-p)^n.
\end{equation}
For arbitrary dependence, the union contains each $E_i$, giving the lower
bound $p$, while Boole's inequality gives the upper bound $np$, clipped at one.
Both bounds are attainable: identical alarm events attain $p$, and disjoint
events attain $np$ when $np\leq1$. Thus the row marginal alone cannot identify
unit risk.

\subsection{Deterministic-activation ceiling}

For a frozen model, $A=f_\theta(O)$ is measurable with respect to $O$.
Conditioning on $O$ makes $A$ fixed, so $Y$ and $A$ are conditionally
independent, and therefore
\begin{equation}
 I(Y;A\mid O)=0.
\end{equation}
Any activation-only decision $h(A)$ is the observable decision
$h\circ f_\theta(O)$. This proof compares information sets and unrestricted
decision classes; it does not imply that a finite text model can learn
$f_\theta$ with the same samples or compute.

\subsection{Distribution-free order statistic and ties}

Let $F_-$ be any negative-unit CDF and let $T_r$ be the $r$th largest of $m$
independent, identically distributed calibration maxima, conditional on the
frozen score function. The deployment unit is an independent draw from the
same distribution. The probability-integral transform
$F_-(Z^-)$ stochastically dominates a uniform random variable, with equality
when $F_-$ is continuous. Consequently the strict tail
$q(T_r)=1-F_-(T_r)=\Pr(Z^-_{\mathrm{new}}>T_r\mid T_r)$ is stochastically
dominated by $\mathrm{Beta}(r,m-r+1)$. Thus, for
$u_r=B^{-1}_{r,m-r+1}(.95)$,
\begin{equation}
 \Pr_{\mathrm{cal}}\{q(T_r)\leq u_r\}\geq .95,
\end{equation}
with equality under continuity. We select the largest $r$ for which
$u_r\leq\alpha$ and alarm only when $S>T_r$. For the four calibration sets,
the resulting UCBs are:
\begin{center}
\begin{tabular}{lrrr}
\toprule
Setting & $m$ & 5\% rank/UCB & 10\% rank/UCB \\
\midrule
MUP & 69 & 1/.042487 & 3/.088444 \\
LASR & 372 & 12/.048471 & 28/.098704 \\
AgentDojo & 100 & 2/.046560 & 5/.089196 \\
ST-Web & 23 & abstain/-- & abstain/-- \\
\bottomrule
\end{tabular}
\end{center}
This strict-exceedance construction remains valid at atoms: a tie makes the
bound conservative because equality does not alarm. The audit records the
selected-value multiplicity, the number of calibration values strictly above
it, and the number of test scores exactly equal to it. All selected primary
values have multiplicity one and every reported test operating point has zero
equality, so replacing the earlier non-strict implementation changes no paper
count. If no rank has $u_r\leq\alpha$, the alarm policy abstains via an infinite
threshold.

\section{LASR Semantic-Identity Sensitivity}

The official LASR test contains 969 semantic request identities. Of these, 207
occur in the official training set and 52 occur in the frozen threshold-
calibration subset. The protocol-declared semantic-unseen mask retains 785 rows
covering 762 identities; it depends only on request identity. Table~\ref{tab:lasr-unseen}
applies the original locked thresholds without refitting or recalibration.

\begin{table}[h]
\caption{LASR results on 762 semantic-unseen requests. Operating rates use the
original strict-exceedance thresholds.}
\label{tab:lasr-unseen}
\scriptsize
\centering
\begin{tabular}{lrrrrr}
\toprule
Monitor & AUROC & 5\% FPR & 5\% TPR & 10\% FPR & 10\% TPR \\
\midrule
Observable & .9066 & .0334 & .6234 & .1054 & .7403 \\
Activation & .9424 & .0180 & .7065 & .0643 & .7662 \\
Joint & .9427 & .0206 & .7065 & .0720 & .7740 \\
\bottomrule
\end{tabular}
\end{table}

At 5\%, Joint-minus-Observable TPR is +.0831 with request-bootstrap 95\% CI
$[.0494,.1169]$; at 10\% it is +.0338 with CI $[.0052,.0623]$. The identity-
disjoint analysis therefore preserves both the activation increment and the
failure to reach the frozen .80 TPR gate.

\section{AgentDojo Exact Counts and Uncertainty}

Table~\ref{tab:adj-ci} reports exact case counts and two-sided 95\%
Clopper--Pearson intervals. These intervals quantify finite test support; they
do not replace the independent calibration guarantee or the paired
case-bootstrap contrast.

\begin{table}[h]
\caption{AgentDojo case operating points: count/denominator and exact 95\% CI.}
\label{tab:adj-ci}
\scriptsize
\centering
\resizebox{\columnwidth}{!}{%
\begin{tabular}{llrr}
\toprule
Budget & Monitor & FPR (75 negative) & TPR (38 positive) \\
\midrule
5\% & Observable & 2/75 [.0032,.0930] & 2/38 [.0064,.1775] \\
 & Activation last & 0/75 [.0000,.0480] & 2/38 [.0064,.1775] \\
 & Joint last & 0/75 [.0000,.0480] & 1/38 [.0007,.1381] \\
\midrule
10\% & Observable & 15/75 [.1165,.3083] & 22/38 [.4082,.7369] \\
 & Activation last & 14/75 [.1060,.2933] & 20/38 [.3582,.6902] \\
 & Joint last & 15/75 [.1165,.3083] & 22/38 [.4082,.7369] \\
\bottomrule
\end{tabular}
}
\end{table}

The paired Joint-minus-Observable 5\% TPR difference is $-.0263$, semantic-case
bootstrap 95\% CI $[-.0789,0]$; at 10\%, the two monitors make identical
positive-case decisions and the paired difference and CI are zero. Because 38
positive cases are below the frozen minimum of 40, these are labeled
support-insufficient and are not used to promote a monitor.

For the prespecified primary Activation-last monitor at the 10\% budget, the
threshold is the fifth-largest of 100 negative calibration maxima. Under a
continuous independent-and-identically-distributed calibration--test null,
the number $K$ of strict test
exceedances among 75 negatives has the rank-predictive law
$K\sim\mathrm{BetaBinomial}(75,5,96)$. Hence
$\Pr(K\geq14)=.001745$. With score atoms, strict exceedance can only make this
upper-tail check conservative. Unlike conditioning on the 95\% UCB, this
calculation integrates threshold-selection uncertainty. It diagnoses
calibration-to-test distribution mismatch; it does not estimate the causal
effect of intervening on an alarm.

\section{Repeated-Attempt Estimand Sensitivity}

The primary AgentDojo result aggregates ten prespecified seeds as admissible
repeated attempts within one semantic security case. This is a sampled
ten-attempt retry-set audit. To test dependence on that operational unit without
generating new behavior, we re-aggregate the frozen scores over nested prefixes
of the original seed order
101, 211, 307, 401, 503, 601, 701, 809, 907, 1009. Table~\ref{tab:retry}
reports selected prefixes; no seed or prefix is chosen by its result.

Formally, evaluation first labels completed attempts, then computes
$Z_G^-=\max_{i:Y_{Gi}=0}S_{Gi}$ and
$Z_G^+=\max_{i:Y_{Gi}=1}S_{Gi}$. The reported class-max FPR and TPR are the
rates of $Z_G^->t$ and $Z_G^+>t$, respectively. A mixed case can enter both
denominators. This estimand measures whether the sampled retry set contains a
safe attempt that would consume review budget and whether it contains an unsafe
attempt that would be detected. It is not a binary any-unsafe case classifier
and does not simulate a controller that halts and suppresses future attempts
after its first alarm; such a controller requires a new sequential contract.

\begin{table}[h]
\caption{Post-hoc AgentDojo retry-policy sensitivity. Entries are realized
FPR/TPR at independently recalibrated FPR-certified thresholds. Support is
negative/positive/mixed semantic cases.}
\label{tab:retry}
\scriptsize
\centering
\resizebox{\columnwidth}{!}{%
\begin{tabular}{llrrrrr}
\toprule
$k$ & Support & \multicolumn{2}{c}{5\% FPR/TPR} & \multicolumn{2}{c}{10\% FPR/TPR} \\
 &  & Observable & Joint & Observable & Joint \\
\midrule
1 & 34/14/0 & .000/.357 & .000/.143 & .088/.643 & .118/.643 \\
3 & 55/31/9 & .036/.194 & .000/.000 & .127/.742 & .109/.710 \\
5 & 66/33/18 & .015/.091 & .000/.030 & .121/.667 & .106/.667 \\
10 & 75/38/31 & .027/.053 & .000/.026 & .200/.579 & .200/.579 \\
\bottomrule
\end{tabular}}
\end{table}

We also evaluate all ten single-seed policies. At 5\%, only four of ten have
enough negative calibration units for a finite threshold. Among those four,
Observable TPR ranges .0476--.3571; Joint-last TPR ranges 0--.1429. At 10\%, all
ten are identifiable, but Observable TPR ranges .0476--.9444 and Joint-last ranges
0--.9565. This variation is not a model comparison: changing the number of
attempts changes both label support and the safe-attempt alarm event. It shows
why a repeated-seed benchmark result cannot be exported to a single-execution
deployment without redefining and recalibrating the alarm policy.

\section{Paired Cross-Benchmark Uncertainty}

Table~\ref{tab:paired-cross} adds paired cluster uncertainty for the activation
increment at each locked operating point. MUP resamples medical specialties;
LASR resamples semantic requests. The intervals are pointwise and use 5,000
replicates with seed 20260713.

\begin{table}[h]
\caption{Paired TPR differences at FPR-certified thresholds.}
\label{tab:paired-cross}
\scriptsize
\centering
\resizebox{\columnwidth}{!}{%
\begin{tabular}{llrr}
\toprule
Setting & Contrast & 5\% difference [95\% CI] & 10\% difference [95\% CI] \\
\midrule
MUP & Joint $-$ Observable & +.2649 [.1529,.4268] & +.0762 [-.0192,.2209] \\
MUP & Activation $-$ Observable & +.1225 [-.0391,.3627] & -.0199 [-.1524,.1698] \\
LASR & Joint $-$ Observable & +.1059 [.0754,.1385] & +.0407 [.0122,.0692] \\
LASR & Activation $-$ Observable & +.1059 [.0754,.1385] & +.0367 [.0082,.0652] \\
\bottomrule
\end{tabular}
}
\end{table}

\clearpage
\balance
\section{Frozen Reproduction Matrices}

\subsection{Eight-field claim-card instantiation}

\begin{table}[H]
\caption{Instantiated Operational Validity Contracts. Cost measurements are
reported as incomplete rather than inferred from predictive performance.}
\label{tab:full-contracts}
\scriptsize
\setlength{\tabcolsep}{2pt}
\begin{tabular}{p{0.12\columnwidth}p{0.24\columnwidth}p{0.31\columnwidth}p{0.23\columnwidth}}
\toprule
Setting & Target / horizon / observable & Identity / unit / comparator / calibration & Cost status \\
\midrule
MUP & high-stakes input; input time; input text & input; one input; frozen lexical observable; negative input maxima & monitor and intervention costs unmeasured \\
LASR & future refusal; pre-response; original request & request hash; semantic request; lexical + three Qwen forecasts; train calibration, official-all plus unseen test & monitor and intervention costs unmeasured \\
AgentDojo & injection execution; pre-action; canonical visible history & suite + user-task + case hash; sampled ten-attempt case; TF--IDF/metadata + three Qwen forecasts; disjoint frozen case calibration/test & hook extraction measured; probe, observer, review, and intervention costs unmeasured \\
ST-Web & task failure; pre-action; browser-visible state & intent-template hash; one template across locked safe/unsafe variants; frozen strong observer + three Qwen forecasts; disjoint template calibration/test & environment reset and rollout costs measured; monitor, review, and intervention costs unmeasured \\
\bottomrule
\end{tabular}
\end{table}

\subsection{Fail-closed claim compiler}

The compiler consumes the typed claim card and already frozen evidence; it has
no interface for fitting, feature selection, threshold search, or subset
selection. Let $E_{\mathrm{spec}}$ require a complete contract and passing
identity audit. Let $E_{\mathrm{rep}}$ require an out-of-sample probe and
held-out ranking evidence. Let $E_{\mathrm{inc}}$ require a declared same-horizon
comparator, point increment at least the frozen minimum, and a positive lower
clustered-confidence bound. For each budget, let $E_{\mathrm{op}}$ require an
identified maximum certifying rank, prospectively prespecified test support,
the budget-local locked TPR minimum $\tau_\alpha$, and realized FPR no larger
than $\alpha$. The maximum claim is
then computed in order:
\begin{equation}
\mathrm{claim}=\begin{cases}
\mathsf{invalid},&\neg E_{\mathrm{spec}},\\
\mathsf{contract},&E_{\mathrm{spec}}\land\neg E_{\mathrm{rep}},\\
\mathsf{representation},&E_{\mathrm{spec}}\land E_{\mathrm{rep}}\land\neg E_{\mathrm{inc}},\\
\mathsf{predictive},&E_{\mathrm{spec}}\land E_{\mathrm{rep}}\land E_{\mathrm{inc}}\land\neg E_{\mathrm{op}},\\
\mathsf{alarm},&E_{\mathrm{spec}}\land E_{\mathrm{rep}}\land E_{\mathrm{inc}}\land E_{\mathrm{op}}.
\end{cases}
\end{equation}
Intervention evidence is an orthogonal verdict: it cannot raise this ladder and
cannot be inferred from an alarm. Missing ranks are permitted only when
rank-one support is analytically insufficient; finite policies must use the
largest certifying rank and a matching Beta UCB. Twenty synthetic
counterexamples exercise these transitions without benchmark outcomes.

\begin{table}[H]
\caption{Compiler outputs from the frozen contracts. ``Rep.'' is representation
evidence; ``Inc.'' is the clustered comparator-increment gate.}
\label{tab:compiler-outputs}
\scriptsize
\setlength{\tabcolsep}{2pt}
\begin{tabular}{p{0.12\columnwidth}p{0.07\columnwidth}p{0.07\columnwidth}p{0.29\columnwidth}p{0.35\columnwidth}}
\toprule
Setting & Rep. & Inc. & Maximum claim & Independent blocker(s) \\
\midrule
MUP & pass & pass & restricted predictive value & test support not prespecified \\
LASR & pass & pass & restricted predictive value & test support not prespecified \\
AgentDojo & pass & fail & representation accessibility & increment, support, transport \\
ST-Web & pass & fail & representation accessibility & increment, calibration/test support \\
\bottomrule
\end{tabular}
\end{table}

\begin{table}[H]
\caption{Parameters frozen before their corresponding evaluation phase.}
\label{tab:repro}
\scriptsize
\begin{tabular}{p{0.23\columnwidth}p{0.68\columnwidth}}
\toprule
Object & Frozen value \\
\midrule
Environment & AgentDojo v1, commit \texttt{089ed468}; official security outcome \\
Agent & Qwen2.5-7B-Instruct; Hermes native tool calls \\
Generation & temperature .7; top-$p$ .9; maximum 512 new tokens \\
Splits & 345 model-fit, 119 calibration, 119 one-use-manifest test cases; user tasks disjoint \\
Seeds & 101, 211, 307, 401, 503, 601, 701, 809, 907, 1009 \\
Boundary & first policy request after the injected tool result \\
Marker & one global constant; no case or seed identity \\
Activation & zero-index decoder layer 14; final token primary, request mean secondary; float32 \\
Observable & exact visible history; character/word TF--IDF; suite/tool metadata; three frozen Qwen-14B forecast log-odds \\
Fitting & shared suite + user-task grouped folds; grouped out-of-fold late fusion \\
Calibration & semantic-case negative maxima; .95 Beta UCB; 5\% and 10\% budgets; strict exceedance; tie/equality audit \\
Test policy & consume one 1,190-identity ten-attempt manifest once; retain failed trajectories; no retraining or threshold changes \\
\bottomrule
\end{tabular}
\end{table}

Model fitting generated 3,450 frozen identities; 3,418 completed and 1,920
boundary-present rows entered feature fitting. Calibration generated 1,190
identities; 1,184 completed and 670 boundary rows across 103 cases were scored.
The one-use test manifest generated all 1,190 identities; 1,165 completed and 602
boundary rows across 82 cases were scored. Prompt reconstruction, token/payload
alignment, constant-marker purity, finite activations, and forecast-template
alignment were 100\% for all eligible rows.

Full SHA256 values for locked scores and independently rechecked metrics are
stored in the machine-readable audit. The supplement provides
manifests and audit commands with portable paths.

\section{Frozen-Score Replay and Screening Transparency}

The empirical replay starts from SHA256-pinned frozen score archives rather
than the manuscript evidence summary. It verifies hashes, reconstructs
semantic-unit maxima, recomputes AUROC, calibration ranks, Beta UCBs,
selected-value ties and equalities, test FPR/TPR, exact intervals, group diagnostics, and
paired bootstraps, and only then compares the recomputed values with the paper.
All frozen-value comparisons pass. It also emits the LASR train/test identity
overlap and semantic-unseen sensitivity; AgentDojo retry sensitivity; and
ST-Web support, modality, worst-modality, and intent-template-bootstrap
diagnostics. It fits no model and exposes no option to change a layer, pooling
rule, threshold budget, template, or subset.

The screening ledger covers all rejected, failed, negative, retrospective,
and clean-rerun branches; the tabled settings are neither the only attempts
nor a random task sample.